# An Error Discovery and Correction for the Family of V-Shaped BPSO Algorithms

Qing Zhao, Chengkui Zhang, Hao Li, Ting Ke

College of Artificial Intelligence, Tianjin University of Science & Technology, Tianjin 300457, China

zhaoqing@tust.edu.cn

**Abstract**

BPSO algorithm is a swarm intelligence optimization algorithm, which has the characteristics of good optimization effect, high efficiency and easy to implement. In recent years, it has been used to optimize a variety of machine learning and deep learning models, such as CNN, LSTM, SVM, etc. But it is easy to fall into local optimum for the lack of exploitation ability. It is found that in the article, which is different from previous studies, The reason for the poor performance is an error existing in their velocity update function, which leads to abnormal and chaotic behavior of particles. This not only makes the algorithm difficult to converge, but also often searches the repeated space. So, traditionally, it has to rely on a low w value in the later stage to force these algorithms to converge, but also makes them quickly lose their search ability and prone to getting trapped in local optima. This article proposes a velocity legacy term correction method for all V-shaped BPSOs. Experimentals based on 0/1 knapsack problems show that it has a significant effect on accuracy and efficiency for all of the 4 commonly used V-Shaped BPSOs. Therefore it is an significant breakthrough in the field of swarm intelligence.

## 1 Introduction

Particle Swarm Optimization (PSO) is one of the most widely used evolutionary algorithms in continuous search spaces, and it is famous for its fast convergence speed and easy programming implementationr[1]. BPSO is the the binary version of PSO algorithm, and widely used in various fields, such as the architecture Search and parameter optimization for Deep Learning Networks[2][3][4], feature selection to enhance the learning algorithm[5][6], Diagnosing Parkinson's Disease[7], Performance improvement of Wireless Sensor Networks[9]. However, It is regrettable that BPSO does not fully inherit the advantages of PSO algorithm in continuous space.

The original BPSO, first proposed by Kennedy and Eberhart[10], has obvious defects in convergence, and difficult to provide satisfactory results. In recent years, various improved versions of BPSO have proposed a variety of better transfer functions. Among these, a kind of V-Shaped transfermation BPSO family, which always has the best optimizaiton effect in the large BPSO family, has aroused wide concern. Most recent improvements [11] rely on adjusting the values of parameters such as w, Vmax and δat different stages of the algorithm to balance exploration and exploitation, but often with poor results. The common problems of these V-Shaped BPSO are: in the early stage, the repeated solution space is often searched, and in the later stage, in order to achieve the convergence effect, search capability often decreases too fast by limiting some parameters, which makes the algorithms easy to fall into local optimum[12].

In this paper, a different view is presented, that the reason why these algorithms have the above problems is that there is a common defect in their speed update formula: the sigmoid transformed velocity v is the probability of bit-value jump, and its value is relative to the current position. When a position jump from 0 to 1 or from 1 to 0 occurs, the probability is not adjusted to the new reference position. Therefore, its value is wrong and can not truly reflect the historical trend of motion. And this error exists in all members of the V-Shaped BPSO family. We will experimentally demonstrate that the existence of this error can lead to abnormal optimization behavior, increase the probability of repeated visits to the explored space, and reduce the rationality of particle behavior. Faced with these "crazy" particles, in order to achieve convergence effect, these algorithms have to rely on a very small value of w in the later stage to reduce the activity of the particles. However, this also makes them quickly lose their search ability, and be easy to fall into local optima.

So that, an improved method of V-Shaped transformation functions based on velocity legacy term correction is proposed in Chapter 3. This kind of new BPSO algorithm family is called VCv-BPSO. It is worth noting that this paper does not propose one improved BPSO, but rather corrects a long-standing error in all current V-shaped BPSO family, which can make these BPSO algorithms more in line with the principles of the original PSO, thereby significantly improving their performance. Subsequently, it is proven that the improved method can achieve significant improvements in both exploration and exploration in all V-shaped BPSOs participating in experiments. On the other hands, the improved method presents a slow and natural convergence, they no longer rely on parameters w, Vmax, or δ to reduce particle activity and achieve convergence, so less likely to fall into local optima. Experimental results have shown that, for low, medium, and high dimensional knapsack problems, the optimal benefits obtained by the improved method can reach 99.8%, 99%, and 98% of the true global optimal solution, respectively. What's even more exciting is that compared to the previous algorithm, the higher the dimensionality of the backpack, the more the percentage of improvement is: in low, medium, and high dimension backpack problems, the average improvement is 3.1%, 8.15%, and 11.95% respectively in the four common V-Shaped BPSOs. And another gratifying phenomenon is the effect of velocity legacy term correction is greater than the effect of different transformation function selection, which has broken the traditional belief about BPSO research..

In order to distinguish whether the algorithms are exploring in new space or jumping back and forth in repeated space, a new evaluation indicator named Effective Haiming Distance Gain is proposed in Chapter 4. Through this indicator , it can be found from experiments that the improved method has strong exploration ability in the early stage, and can effectively avoid the back and forth oscillation in the repeated space, which is the main reason for its higher efficiency. And in the later stage, the search ability of the improved algorithm decreases slowly and presents a natural and smooth convergence state, which is also the reason why the improved it is less likely to fall into local optimum than the algorithm before improvement. At the same time, it also show that why the existing BPSO variants have higher and higher jumping probability and have to rely on the parameter adjustment to convergent is just because of the error in the velocity update formula.

## 2 Related Work

In the past few years, a variety of improved methods have been proposed for the shortcomings of the original BPSO. In response to the common problem of lack of exploitation ability in the later stage, some studies have adopted a composite algorithm approach combined with other algorithms, such as the combination of binary black hole algorithm and BPSO in [13], which to some extent avoids local minima. Reference [14] replaces the worst-case particle trapped in local optima by introducing catfish particles. Reference [15] combines BPSO with adversarial learning, chaotic mapping, fitness based dynamic inertia weights, and mutation for feature selection in text clustering, achieving high clustering accuracy and improving the performance of BPSO. [16] issued a PSO-based algorithm that hybridized Particle Swarm Optimization (PSO) and Hill Climbing (HC) which is applied to high school timetabling problem.

But these methods have not fundamentally solved the fundamental problem why PSO algorithm's performance is not satisfactory when applied to binary space. Moreover, these composite algorithms are often more complex, and the traditional high-efficiency advantage of PSO are difficult to continue in the binary space。

In addition to these composite variant algorithms, other studies are focused on the conversion functions of BPSO, as these researchers generally believe that the use of inappropriate transfer functions is the main reason for the poor performance. A lot of research has been done on this problem to balance exploration and exploitation. For example, in [11] , a time-varying transfer function has been issued in order to achieve higher exploration in early stage and higher exploitation in late stage. And many other BPSO variants have been developed to tackle this issue such as [17, 18, 19, 20]. However, a good balance between exploration and exploitation cannot be easily achieved. These studies believe that the failure of BPSO is due to their strong exploration and weak exploitation. After analyzing by the new indicators for detecting search ability proposed in this article, we can see that these BPSO algorithms are just an illusion presented by particles frequently jumping back and forth in the repeating space. The effective exploration ability of these algorithms is actually insufficient, and the motion of particles is lack of rationality and targetedness. In order to converge, it is necessary to limit the values of some parameters to reduce the particle's activity. However, this approach further makes it very easy to fall into local optima in the later stage. Therefore, simply adjusting some parameters to achieve a balance between exploration and exploitation cannot solve the fundamental problem.

Other studies are dedicated to improve the speed definition and speed update formula of BPSO. These studies suggest that it may not be appropriate for the velocity formula of BPSO to be exactly the same as that of PSO. As in binary space, particle motion is not as smooth as in continuous space, with only two possible values 0 and 1. Therefore, the idea of moving in one direction in a binary search space is not applicable. Therefore, in order to better guide particles and avoid premature convergence, some methods have been proposed. In [21], multiply the velocity by a random number, which aims to prevent the premature convergence in BPSO. But the improvement is limited and has not been comprehensively compared with the original BPSO. In [22], the concept of flipping probability is used instead of the original velocity. The experimental results show that it outperforms the standard BPSO in both effectiveness and the efficiency in feature selection problems. In [23], a new momentum concept of BPSO based on probability vectors was proposed, in two well-known binary problems: knapsack and feature selection. In [24]，a mutation operator is used to improve the diversity of the swarm, but in comparison with BPSO, this method are sometimes more expensive and have a problem of velocity and previous position ignorance.

Although these algorithms have solved some of the problems of the original BPSO, they still cannot fully reflect the advantages of the original PSO - high efficiency, directness, and good optimization results. They either lack inheritance of historical location and speed, or their optimization ability is not significantly improved, or their efficiency has decreased. But the improvement of the speed meaning and speed update formula that they focus on is indeed the key to improving BPSOs. After analysis, the fundamental reason for the poor performance of BPSO is that some parameters in the velocity update formula have undergone physical changes from PSO in continuous space to BPSO in discrete space. And these parameters continue to be applied directly in the original way, so that the algorithm no longer fully conforms to the physical meaning of PSO, and can no longer guide particles to search for optimization rationally. This is also the main reason why the advantages of the original PSO algorithm cannot continue in binary discrete space.

Therefore, this article will correct the parts in the current BPSO algorithm that do not conform to the physical meanings of PSO. The improvement method is concise and efficient, thus fundamentally solving the problem of weak optimization ability of BPSO. [25] conducted an experimental comparison between four mainstream V-shaped BPSO algorithms and four mainstream S-shaped BPSO algorithms. The results showed that the V-shaped BPSO family has advantages over other BPSO variants in avoiding local minima and improving convergence speed. Therefore, this article focuses on the V-shaped BPSO family, which has the best performance in BPSO field. We have discovered the inherent physical errors in its speed update formula, and through correction, it has regained the good optimization ability and search efficiency of the original PSO algorithm. It no longer relies on complex parameter tuning and the assistance of other intelligent algorithms. The improved method is direct, efficient, and easy to implement, allowing the excellent performance of the PSO algorithm to continue to be realized in binary discrete space.

## 3 New BPSO Algorithms Based on Velocity Legacy Term Correction, VCv-BPSO

### 3.1 Arithmetic Principle

In this section, it will be seen that after correcting the velocity legacy term, the historical move tendency of particles can be correctly continued, and subsequent searches can consistently adhere to their goals, so that the particles' behavior can be more intelligent and rational.

For the V-Shaped BPSO family, the velocity after transformation of the sigmoid function is the jump probability of the bit value, which is relative to the current position. When a bit value jump occurs, the reference position changes, and the expression of the velocity legacy term should be relative to the new reference position. Therefore, its value needs to be corrected in order to truly express the historical accumulated motion trend.

$$\mathrm{v}_{\mathrm{id}} = \omega \cdot v_{id} + c_1 \cdot rand() \cdot (p_{id} - x_{id}) + c_2 \cdot rand() \cdot (p_{gd} - x_{id}) \qquad (1)$$

In formula (1), the part $c_1 \cdot rand() \cdot (p_{id} - x_{id}) + c_2 \cdot rand() \cdot (p_{gd} - x_{id})$ represents the current optimization trend, and the part $\omega \cdot v_{id}$ is the historical search trend which we refer to as the velocity legacy term in this article. The velocity legacy term represents the optimization trend left over from the previous steps. its positive value represents a trend towards 1, while its negative value represents a trend towards 0, and the larger the absolute value, the more obvious and firm the trend of this movement. To illustrate the necessity of correcting the velocity legacy term when a position jump occurs, we consider the following scenario:

Assuming the current speed is $\mathrm{V}_{\mathrm{id}}$ and the current position is 0, the value of sigm( $\mathrm{V}_{\mathrm{id}}$ )

represents the jumping probability, as the jump function is:

$$x_{i+1} = \begin{cases} 1 - x_i, & \text{if } rand() < sigm(v_{id}) \\ x_i, & \text{otherwise} \end{cases} \tag{2}$$

Note that the jumping probability at this point is actually the probability of jumping to 1 relative to position 0. Assuming that a jump has indeed occurred this time, so the next position is changed to 1. If the velocity $v_{id}$ is not corrected, the historical trend is still the probability of jumping to 1. According to Formula (2), the higher its value, the greater the probability of jumping back to 0. In fact, upon careful analysis, it can be found that this has already violated the principles of the PSO algorithm. In this example, the historical trend legacy should still be the probability of turning to 1, rather than the probability of jumping back to 0, because in the previous optimization process, it was found that turning to 1 is more likely to obtain a better solution, and the expression of this trend should be stable. The reason for the instability here is precisely because for BPSO, unlike PSO, there are only two extreme positions 0 and 1. When the position changes, the reference position fundamentally changes. If the original probability of turning 1 is not corrected, it will become the probability of continuing to jump and turning 0. This is obviously not what we want to express, so it is necessary to correct the velocity legacy term here. Subsequent analysis will further reveal that the fundamental reason for the unsatisfactory performance of various variants of BPSOs is the lack of correction for the velocity legacy term when the bit value jumps.

For how to correct it, we can consider it this way: sigma ( $v_{id}'$ ) should be equal to 1-sigma ( $v_{id}$ ), because at this point, position 1 is taken as the reference position, and the remaining velocity term in the previous step represents the probability of maintaining 1 as sigma ( $v_{id}$ ), while the probability of jumping back to 0 is 1-sigma ( $v_{id}$ ). Therefore, we should make: sigm(vid’ )=1-sigm(vid)，and derived the corrected value of vid' from it. This correction method for the velocity legacy term is suitable for all V-Shaped BPSO algorithms.

In summary, the correction method for the velocity legacy term is as follows: if a position jump occurs during a certain time, in the next velocity update formula, the previous velocity $v_{id}$ should be corrected to $v_{id}'$ , that is:

$$v_{id} = \omega \cdot v_{id}' + c_1 \cdot rand() \cdot (p_{id} - x_{id}) + c_2 \cdot rand() \cdot (p_{gd} - x_{id}) \tag{3}$$

And the calculation method for $v_{id}'$ is: Let $v_{id}'$ satisfy:

$$sigm(v_{id}') = 1 - sigm(v_{id}) \tag{4}$$

Thus deriving the expression for $v_{id}'$ .

Here we take one of the V-Shaped BPSO algorithms as an example, which has a conversion function of $sigm(v)=\frac{v^2}{1+v^2}$. If a position jump does occur after generating a random number and substituting it into the following equation:

$$x_{i+1}=\begin{cases}1-x_i, & \text{if } rand()<sigm(v_{id})\\ x_i, & \text{otherwise}\end{cases} \tag{5}$$

Let $v_{id}'$ satisfy the following equation:

$$sigm(v_{id}')=\frac{v_{id}'^2}{1+v_{id}'^2}=1-sigm(v_{id})$$

$$=> \quad \frac{v_{id}'^2}{1+v_{id}'^2}=1-\frac{v_{id}^2}{1+v_{id}^2}=\frac{1}{1+v_{id}^2}$$

$$=> \quad v_{id}'=\frac{1}{v_{id}} \tag{6}$$

Therefore, for this V-Shaped BPSO algorithm, when a particle undergoes a position jump, its velocity legacy term should be modified according to equation (6).

## 3.2 Four common V-Shaped BPSOs and their velocity correction methods

The commonly used V-Shaped BPSO algorithms mainly have the following sigmoid function forms, which are referred to as VT1, VT2, VT3, and VT4 in this article, as shown in equations(6)-(9), and their images are shown in Figure 1.

$$VT1:\ sigm(v_{id})=\left|\frac{2}{\pi}\arctan\left(\frac{\pi}{2}v_{id}\right)\right| \tag{7}$$

$$VT2:\ sigm(v_{id})=\frac{v_{id}^2}{1+v_{id}^2} \tag{8}$$

$$VT3:\ sigm(v_{id})=\left|\tanh(v_{id})\right| \tag{9}$$

$$VT4:\ sigm(v_{id})=\frac{2}{1+e^{-|v_{id}|}}-1 \tag{10}$$

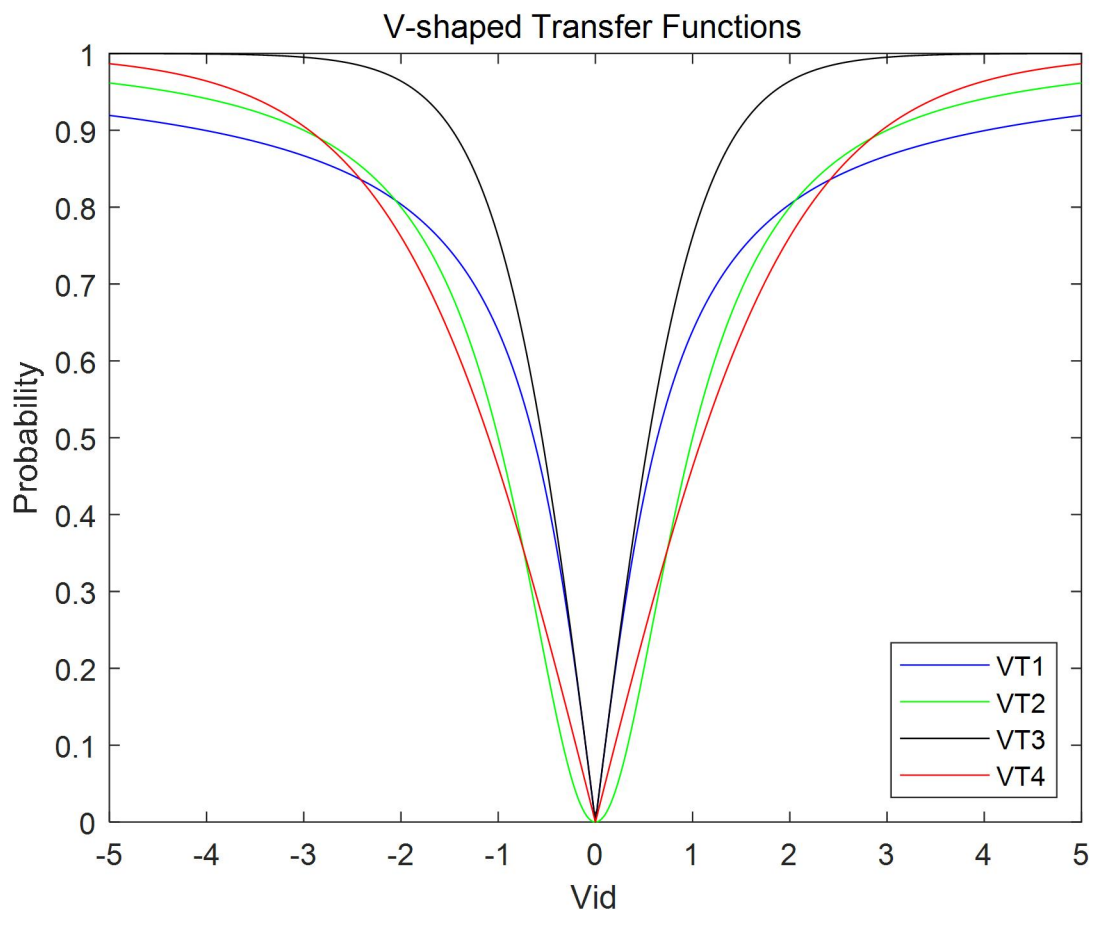

Fig. 1 four common transfer functions for V-Shaped BPSOs

According to the correction formula (4), the speed correction functions for the four V-shaped BPSO algorithms can be derived, resulting in four new BPSO algorithms, denoted as $VC_{v1}-BPSO$, $VC_{v2}-BPSO$, $VC_{v3}-BPSO$, and $VC_{v4}-BPSO$. The derivation results are shown in Table 1. The specific derivation process of $VC_{v1}-BPSO$, $VC_{v3}-BPSO$, and $VC_{v4}-BPSO$ can be found in Appendix.

Table 1 Velocity correction function for the four common V-Shpaed BPSOs

| The original V-shaped BPSOs | Sigmoid transfer function | New BPSOs based on velocity correction | velocity correction function |
|---|---|---|---|
| $V_{T1}$-BPSO | $sigm(v_{id}) = \left\|\frac{2}{\pi}\arctan(\frac{\pi}{2}v_{id})\right\|$ | $VC_{v1}$-BPSO | $v_{id}' = \frac{4}{\pi^2 \cdot v_{id}}$ |
| $V_{T2}$-BPSO | $sigm(v_{id}) = \frac{v_{id}^2}{1+v_{id}^2}$ | $VC_{v2}$-BPSO | $v_{id}' = \frac{1}{v_{id}}$ |
| $V_{T3}$-BPSO | $sigm(v_{id}) = \left\|\tanh(v_{id})\right\|$ | $VC_{v3}$-BPSO | $v_{id}' = \begin{cases} \frac{1}{2}\ln\frac{e^{v}+3e^{-v}}{e^{v}-e^{-v}}, & v \geq 0 \\ \frac{1}{2}\ln\frac{e^{-v}-e^{v}}{3e^{v}+e^{-v}}, & v < 0 \end{cases}$ |
| $V_{T4}$-BPSO | $sigm(v_{id}) = \frac{2}{1+e^{-\|v_{id}\|}} - 1$ | $VC_{v4}$-BPSO | $v_{id}' = \begin{cases} \ln\frac{1+3e^{-v_{id}}}{1-e^{-v_{id}}}, & v_{id} \geq 0 \\ \ln\frac{e^{-v_{id}}-1}{3+e^{-v_{id}}}, & v_{id} < 0 \end{cases}$ |

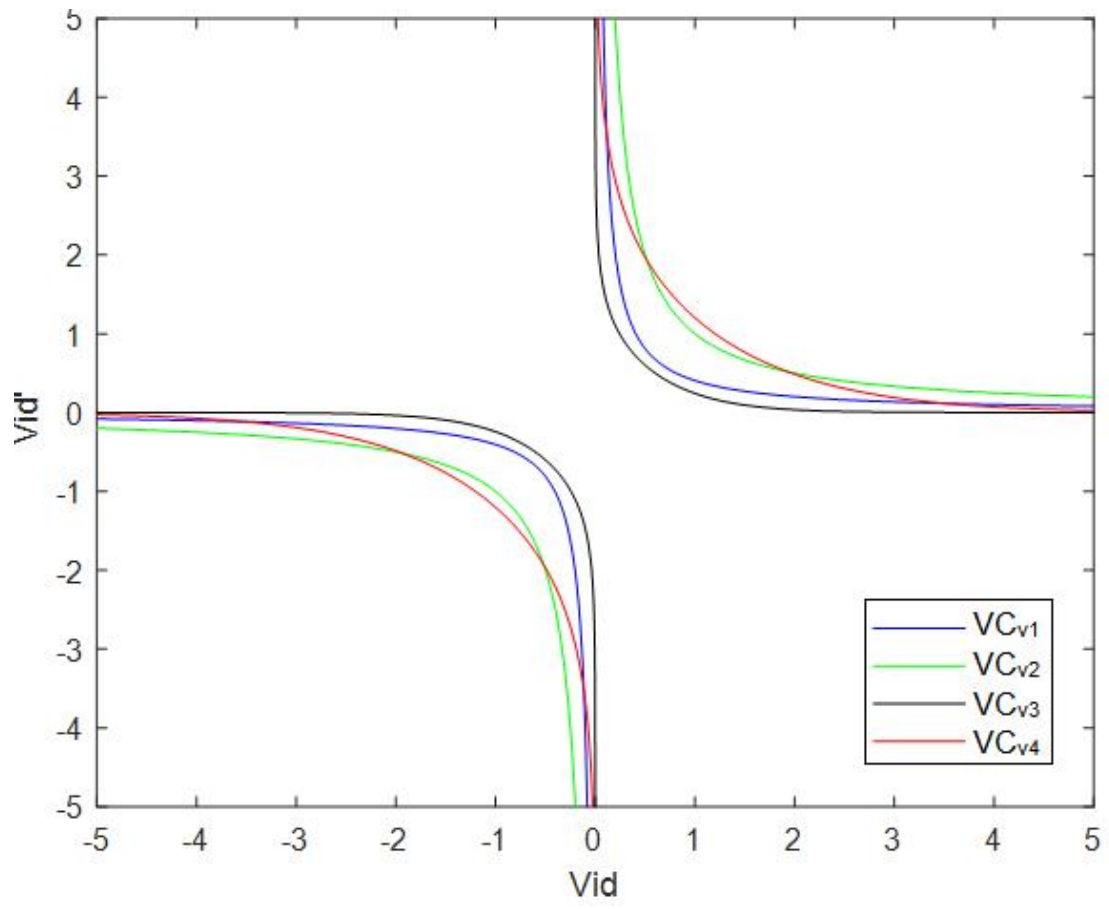

Fig.2 Image of velocity correction functions corresponding to four V-Shaped BPSO algorithms

From Figure 2, it can be seen that, like the four original conversion functions, the four speed correction functions also have similar images.

## 4. A new measurement for the exploration and exploitation ability - Effective Hamming Distance Gain

In order to better analyze the effectiveness of the velocity legacy term correction method proposed, a new indicator for evaluating particle exploration and exploitation ability is proposed.

Compared to traditional measurement methods, it can better distinguish whether the BPSO algorithm performs efficient search in new spaces or pretends to work hard in repeated spaces.

The problem with the velocity legacy term in the original V-shaped function BPSO algorithm and its various variants can easily lead to particles oscillating back and forth between repeated or similar positions. This oscillation can obtain a high exploration ability score according to the measurement method based on the Hamming distance[26] between two consecutive iterations proposed by Md. Jakirul Islam[11], but its true search ability is relatively limited. However, this traditional measurement method cannot accurately distinguish it.

### 4.1 Measurement method of Md. Jakirul Islam

In [11], it is believed that in BPSO, a large absolute velocity leads to exploitation, and a small absolute velocity leads to exploration. Therefore, The exploration and exploitation phase can be determined for the binary search space by measuring the hamming distance between the previous position xki and the current position xki +1 of the i-th particle of BPSO. The equation for measuring the hamming distance is:

$$dist_i = \sum_{j=1}^{d} |x_{i,j}^{k} - x_{i,j}^{k+1}|. \tag{11}$$

However, in fact, if a particle oscillates back and forth in two (or a few) local spaces, even if its velocity measured by Hamming distance remains high, its true and effective exploration and explosion ability is not strong, and it still searches in a very limited local space, as shown in Figure 3. Therefore, a more suitable method for measuring the search ability of particles is proposed.

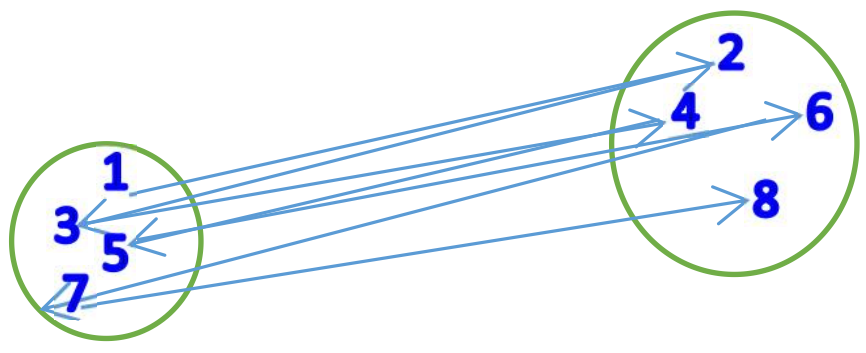

Fig.3 The situation where particle Hamming velocity is high but exploration ability is limited

### 4.2 A new measurement method for exploration - Effective Hamming Distance Gain

4.2.1 Effective Hamming Distance Gain

This article argues that in the process of iteration, particles can only perform effective spatial search if they find a location different from the one they have already searched for. The farther the new location is from the previously searched location, the stronger its exploration ability. Otherwise, even if the particle's speed is high between two consecutive iterations, it still does not have strong exploration ability. Therefore, after each iteration, we calculate the Hamming distance between the particle's new position and its closest historical position, which we refer to as the Effective Hamming Distance Gain of this iteration and record as Dist_ Eff.

$$Dist_eff_i^k = \min_{1 \le r \le k-1} \left( \sum_{j=1}^{d} \left| x_{i,j}^k - x_{i,j}^r \right| \right) \tag{12}$$

Here, d represents the dimension of the particle, and $x_{i,j}^{r}$ represents the value of the jth position of particle i after the r-th iteration. According to (12), if the effective distance gain $Dist_eff_i^k$ is found to be sufficiently large, then it can be said that BPSO is exploring the search space in this iteration, otherwise, it is exploiting the search space.

Consider the differences between the two measurement methods in the three scenarios shown in Figure 4. For simplicity, the particles in the example have only 4 dimensions. Considering the increase in spatial search ability from iteration 1 to iteration 4, the measurement method of Md. Jakarta Islam and the effective Hamming distance gain proposed in this article are used to measure the increase, denoted as $Dist_{1-4}$ and $Dist_eff_{1-4}$, respectively.

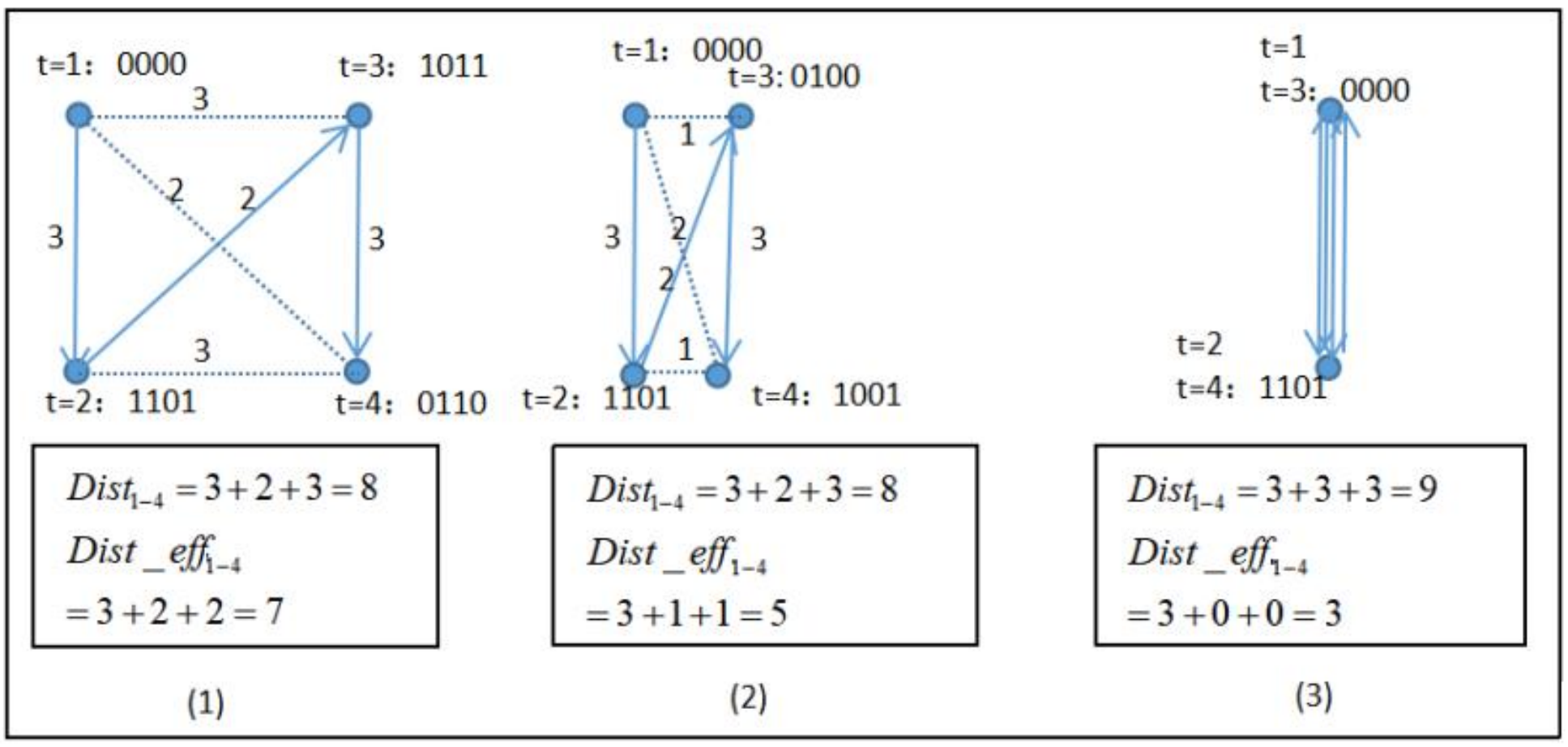

Fig. 4 comparison of the two measurement methods in three typical scenarios

In this figure, the line segment connecting two positions is marked with the Hamming distance between the two positions, and the solid line represents the continuous jumping action of the particle. In the three cases shown in the figure, the results calculated according to the measurement method of Md. Jakirul Islam are basically consistent, with a=a in the first two cases and b=b in the third case. But it is obvious that in reality, in the first case, the four positions of the particles are more dispersed, while in the third case, they only oscillate back and forth between the two positions. In the second case, it is between the first and third cases, and the positions of the particles are similar at times 1 and 3, and at times 2 and 4. According to the measurement method of Md. Jakarta Islam, the Dist1-4 in the third case is actually larger. Therefore, it can be seen that the measurement method of Md. Jakarta Islam cannot fully measure the particle's exploration ability, as it cannot distinguish back and forth oscillations between repeated or similar positions. According to the method based on effective Hamming distance gain proposed in this article, the calculated results precisely reflect the actual rate of divergence of the search space in situations 1, 2, and 3.

If we accumulate the effective Hamming distance gains generated by each iteration over a period of time, we can understand the strength of exploration and exploitation capabilities during this period.

4.2.2 Measurement of particle Useless jump volume

However, it should be noted that the original measurement method based on Hamming distance is not meaningless. By calculating the total difference between the Hamming distances Dist and Dist_eff for each iteration over a period of time, it is possible to understand how many invalid jumps in the repeated space the particle has experienced during this period. Here, define the particle useless jump volume ( PUJV ) over a period of time:

$$\mathrm{PUJV}_m^n = \sum_{i=1}^{ParticalNum} \sum_{k=m}^{n} (Dist_i^k - Dist_eff_i^k) \tag{13}$$

This measurement index can be used to analyze the reason of the high efficiency of the improved BPSO algorithm in the experiment.

## 5 Experiment Results and Analysis

### 5.1 The 0-1 knapsack problem

The 0-1 knapsack problem is selected to test the the effectiveness of the proposed improved method, and it can be described using the following example. Suppose we have *n* items and each item *i* has a weight $w_i$ and a profifit $p_i$ , where $i = 1, 2, . . . , n$. The goal is to select a subset of items in order to maximize the total profit such that the total weight of the selected items does not exceed the knapsack capacity *C*. To model this problem, $x_i$ is introduced as a decision variable for each item. If the *i*-th item is selected then the value of $x_i$ is set to 1, otherwise its value is set to 0. Formally, the 0-1 knapsack problem can be expressed as:

$$\begin{aligned} &\text{maximize} && \sum_{i=1}^{n} p_i x_i \\ &\text{subject to} && \sum_{i=1}^{n} w_i x_i \le C, \\ &\textit{where} && x_i \in \{0,1\},\ i = 1, \ldots, n. \end{aligned}$$

These instances are generated randomly, the optimal solutions of these instances are calculated by dynamic programming method. For the sake of comprehensiveness, mainly four types of test instances have been identified, by taking into account the correlation between the profits and weights of the items [11]: uncorrelated instances, weakly correlated instances, strongly correlated instances, and inversely strongly correlated instances. According to [11], the generation procedure of these four types is provided below by considering a variable R ∈ N: R is a positive number.

• Uncorrelated instance (UCI): For this type, the weight and profit of an item i are selected as: wi ∈ U[1, R] and pi ∈ U[1, R], respectively.

• Weakly correlated instance (WCI): For this type, the weight of the i-th item is specified as: wi ∈ U[1, R] and the profifit of that item is specified as: pi ∈ U[wi − R/10, wi + R/10] subject to pi ≥ 1.

• Strongly correlated instance (SCI): The weight and profifit of the i-th item of the SCI instances are selected as: wi ∈ U[1, R] and pi =wi + R/10, respectively.

For each instance type, the knapsack capacity C is set to a certain percentage (S) of the sum of weights as:

$$C = S \times \sum_{i=1}^{n} w_i,$$

where the value for R is set to 1000, S is set to 0.5, and the value of the dimension of the backpacks is set to 100, 500 and 1000, respectively.

## 5.2 Comparison of algorithms before and after improvement for low dimensional knapsack

The first experiment is used to compare $VC_{v1}$-BPSO 、 $VC_{v2}$-BPSO 、 $VC_{v3}$-BPSO and $VC_{v4}$-BPSO with $V_{T1}$-BPSO 、 $V_{T2}$-BPSO 、 $V_{T3}$-BPSO and $V_{T4}$-BPSO respectively over the low-dimensional ( D =100) 0-1 knapsack instances, so as to verify the effectiveness of the velocity legacy term correction algorithm proposed in this paper.

5.2.1 Comparison of maximum profit value found in search

Firstly, determine the optimal w value for each algorithm to obtain the optimal profit value through experiments. Each algorithm is executed 20 times at different w values, and the average value is calculated. After testing, it has been found that most original algorithms without corrected velocity legacy terms use variable parameter w to achieve the best results. That is, starting with a larger value of w to make the search have strong exploration ability, and later reducing it to a smaller value to ensure good convergence. The improved algorithm after correcting the velocity legacy term basically does not need to rely on smaller w values to ensure convergence. We compare the optimal profit values obtained by each V-Shaped BPSO algorithm before and after speed correction under the main w parameters. For low dimensional backpacks, due to the randomness of UCI, we first select UCI for experimentation. In the experiment, d=100, m=20, c1=c2=2 were taken, and before correction, vmax=- vmin=5. After correction, no upper or lower speed limits were set, and the iteration limits were all 1000 rounds. The specific experimental results are shown in Table 4. The percentage in the table represents the ratio of the optimal profit value to the true global optimal solution calculated by the dynamic programming algorithm.

Table 2 Experimental results of low dimensional knapsack

| $V_T$ | W=0.4 | W=0.6 | W= 0.9-0.4 | W= 1.0-0.4 | $VC_v$ | W=0.99 | W=1.0 | W=1.2-0.99 | W=1.2-1.0 |
|---|---|---|---|---|---|---|---|---|---|
| $V_{T1}$ | 36262.35 ( 92.2%) | **38606.35 ( 98.2%)** | 38421.60 ( 97.7%) | 38366.35 ( 97.5% ) | $VC_{v1}$ | 39174.30 (99.6%) | **39263.40 ( 99.8% )** | 38645.90 ( 98.3% ) | 38139.15 ( 97.0% ) |
| $V_{T2}$ | 33306.00 (84.7% ) | 36244.75 ( 92.2%) | 37693.85 ( 95.8% ) | **37695.30 ( 95.8%)** | $VC_{v2}$ | 38724.00 ( 98.5% ) | **39118.85 ( 99.5% )** | 39039.45 ( 99.3% ) | 38758.00 ( 98.5% ) |
| $V_{T3}$ | 35842.70 ( 91.1% ) | not convergent | 37680.50 ( 95.8% ) | **37680.70 ( 95.8% )** | $VC_{v3}$ | 37892.95 ( 96.3% ) | 38187.25 ( 97.1% ) | **39243.95 ( 99.8% )** | 39215.55 ( 99.7% ) |
| $V_{T4}$ | 35136.95 ( 89.3% ) | 37865.85 (96.3% ) | 38064.00 ( 96.8% ) | **38198.20 ( 97.1% )** | $VC_{v4}$ | 38710.30 ( 98.4% ) | 38956.10 ( 99.0% ) | **39269.90 ( 99.8% )** | 39199.90 ( 99.7% ) |

From the above comparison results, it can be seen that each VC-BPSO algorithm with speed correction has achieved significant improvement compared to the original algorithm. Its average maximum value is very close to the true global optimal solution obtained by the dynamic programming algorithm. Except for VCv2-BPSO, the other three have reached 99.8%, while VCv2-BPSO can also reach 99.5% after correction. From this, it can be seen that regardless of which of the four typical V-Shaped transformation functions is selected, the velocity legacy term correction algorithm proposed in this paper is very effective. For conversion functions 2 and 3, the improvement effect is most significant, with a 4.1% and 3.8% increase in profit values compared to the results under the optimal parameters before correction. The average improvement rate of the speed legacy term correction on the four conversion functions is 3.1%.

Moreover, from the above table, we can also observe that before improvement, there was a significant difference in the optimal profit values obtained by the four conversion functions. However, after adjusting for the velocity legacy term, the differences among the four conversion functions decreased, and the final optimal profit values tended to be consistent, all ranging from 99.5% to 99.8%. The result is not only much better than the algorithm before improvement, but also the effect of speed legacy term correction seems to be greater than the selection effect of different transformation functions.

5.2.2 Execution Efficiency

After further analysis, this article found that the correction of the velocity legacy term not only increased the maximum profit value, but also achieved a significant improvement in execution efficiency. Here, we select the w parameter values to achieve optimal performance for each algorithm in Table 2 to compare the number of iterations and execution efficiency, as shown in the Figure 5 and Table 3.

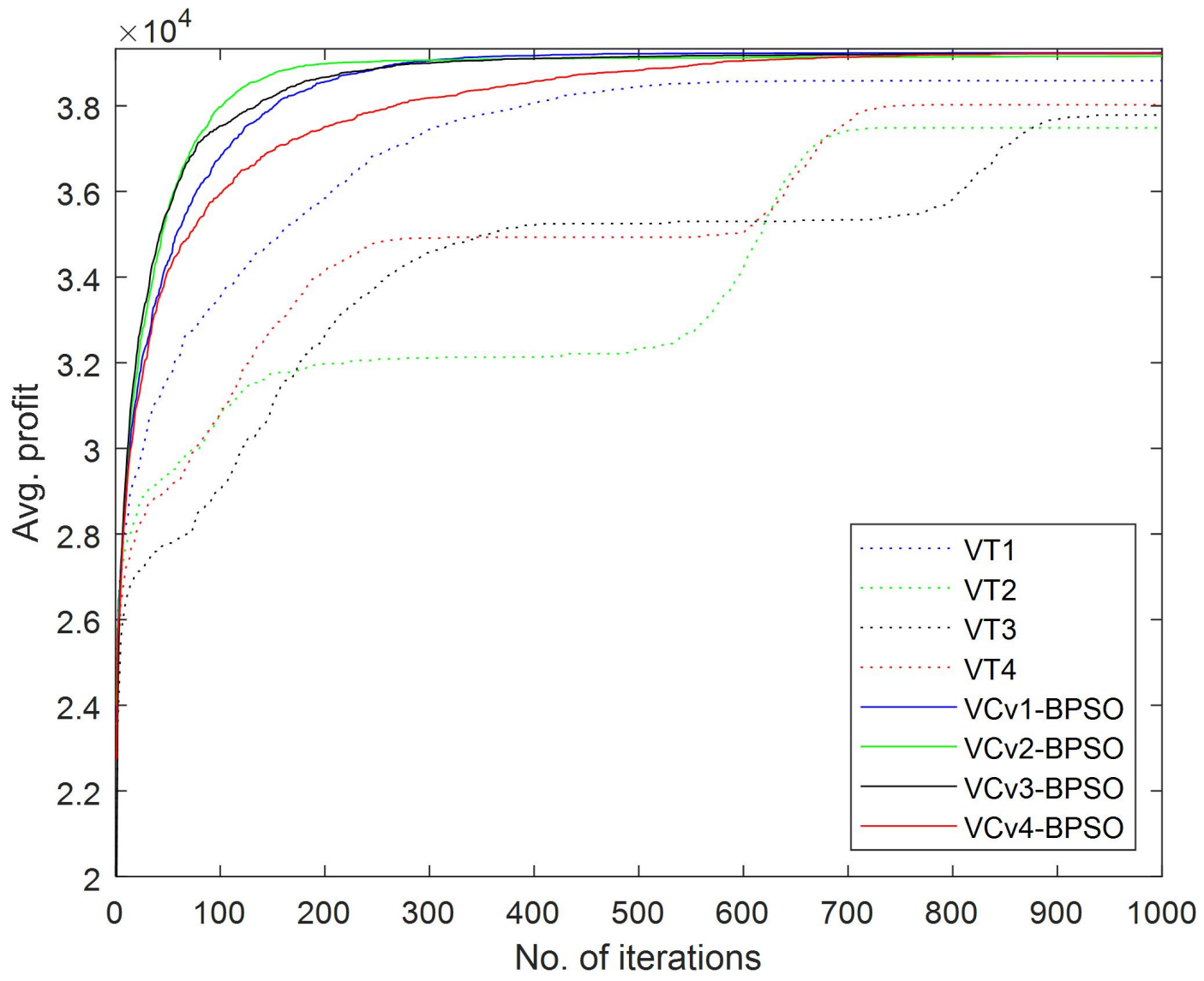

Fig. 5 Comparison of execution efficiency before and after velocity correction

From the Figure 5, it can be seen that the improved algorithm can quickly search for good solution in the early stage of iterations and quickly enter a gradually converging state. The overall convergence speed is significantly better than that before the improvement. This further proves that by correcting the velocity legacy term, search behavior becomes more rational, and particles can approach the optimal target more quickly.

Through Figure 5, we also found that the various methods of the modified VC-BPSO family performed very well in the early stage, and could quickly shorten the distance to the optimal solution. However, in the later stage, the average complete convergence time of the four improved algorithms was only slightly better than that of the four algorithms before correction, as shown in Table 3 below. On average, the four algorithms shortened the number of convergence rounds by 12.6%.

In fact, the improved method actually discovers the final optimal solution at an early stage, but occasionally jumps out later on. If comparing the number of rounds in which the four algorithms first search for the final optimal solution before and after improvement, as shown in Table 4, the four algorithms after improvement have an average reduction of 27.8% compared to before improvement. After analysis, we found that the repeated oscillations in the later stage are caused by the w value being greater than or equal to 1. By fine-tuning the algorithm's w value in the later stage, convergence efficiency can be further improved, and only a small amount of profit value will be sacrificed. Its optimization ability is still far greater than that of the algorithm before correcting the velocity legacy term. The specific experimental results will be seen in the next section when analyzing the impact of w value in detail.

If comparing the four conversion functions, in terms of convergence speed, VCv2-BPSO is the fastest and VCv4-BPSO is the slowest. However, from the specific performance in Table 2, VCv4-BPSO can also achieve the highest optimal profit value, while VCv2-BPSO has the lowest. Therefore, it can be considered that the slow decrease in exploitation in the later stages of iterations can indeed make the algorithm less likely to fall into local optima.

Table 3 Comparison of the average number of convergence rounds

| $V_T$ | Average convergence round | $VC_v$ | Average convergence round |
|---|---|---|---|
| $V_{T1}$ | 588.20 | $VC_{v1}$ | 695.9 |
| $V_{T2}$ | 597.17 | $VC_{v2}$ | 272.6 |
| $V_{T3}$ | 900.93 | $VC_{v3}$ | 681.33 |
| $V_{T4}$ | 729.97 | $VC_{v4}$ | 810.47 |

Table 4 Comparison of average number of rounds for the first discovery of the final optimal solution

| $V_T$ | Average number of rounds for the first discovery of the final optimal solution | $VC_v$ | Average number of rounds for the first discovery of the final optimal solution |
|---|---|---|---|
| $V_{T1}$ | 521.90 | $VC_{v1}$ | 455.13 |
| $V_{T2}$ | 592.70 | $VC_{v2}$ | 265.23 |
| $V_{T3}$ | 897.47 | $VC_{v3}$ | 543.17 |
| $V_{T4}$ | 725.77 | $VC_{v4}$ | 713.10 |

5.2.3 Fine tuning of w in the later stage

Through experiments, it was found that although the improved BPSO algorithm can achieve smooth and natural convergence without relying on low w values, adjusting the w value appropriately in the later stage can significantly improve the convergence speed. However, it only has a slight impact on the optimal profit value and is still far better than the BPSO algorithm without speed correction.

We conducted experiments on four improved algorithms, VCv1-BPSO, VCv2-BPSO, VCv3-BPSO, and VCv4-BPSO, based on their respective optimal w parameters in Section 5.2.1. We also fine tuned the w values downwards by 0.01~0.1 and compared them with the algorithms before speed correction. The results are shown in Figure 6 and Table 5. The average number of Useless jump volume in Table 5 is calculated using the formula in Section 4.2.2.

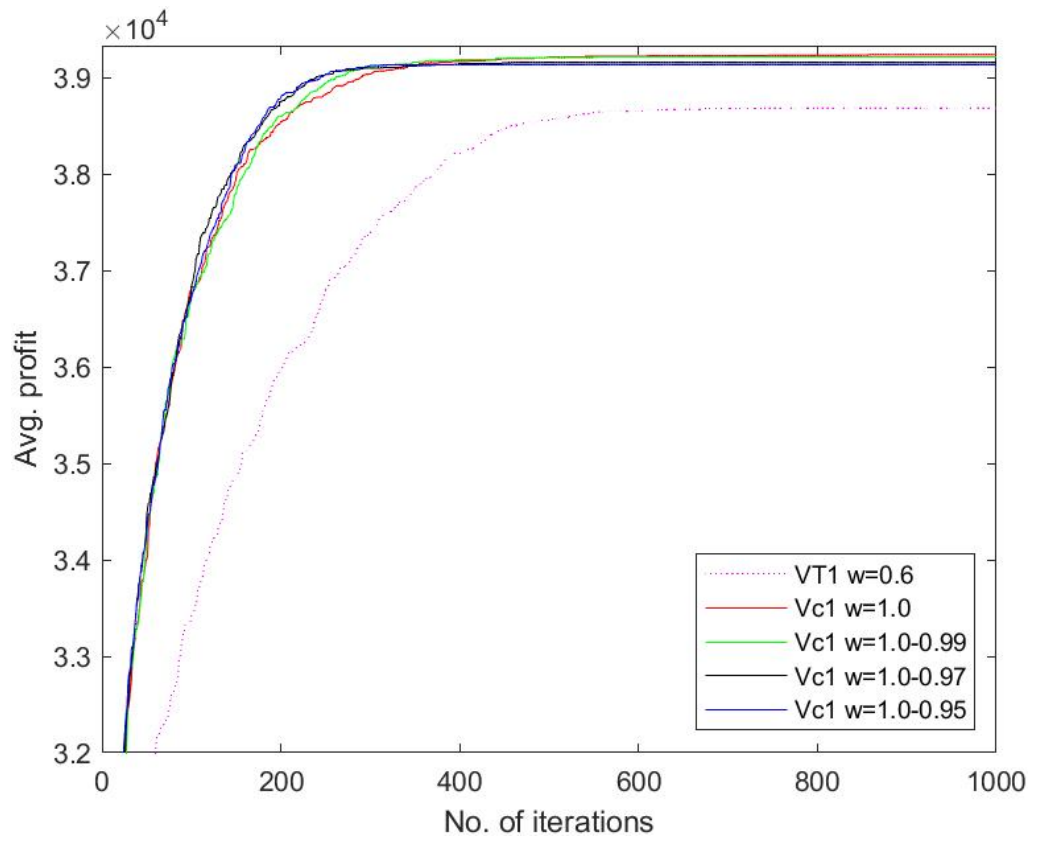

(a) $VC_{V1}$

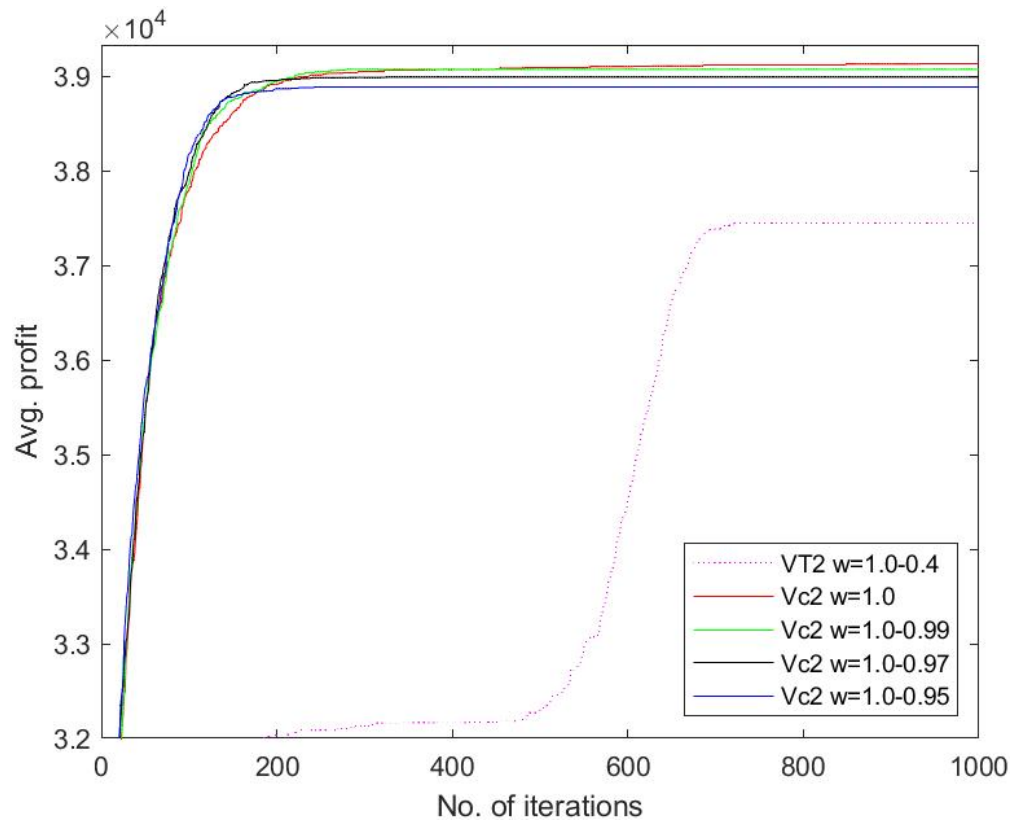

(b) $VC_{V2}$

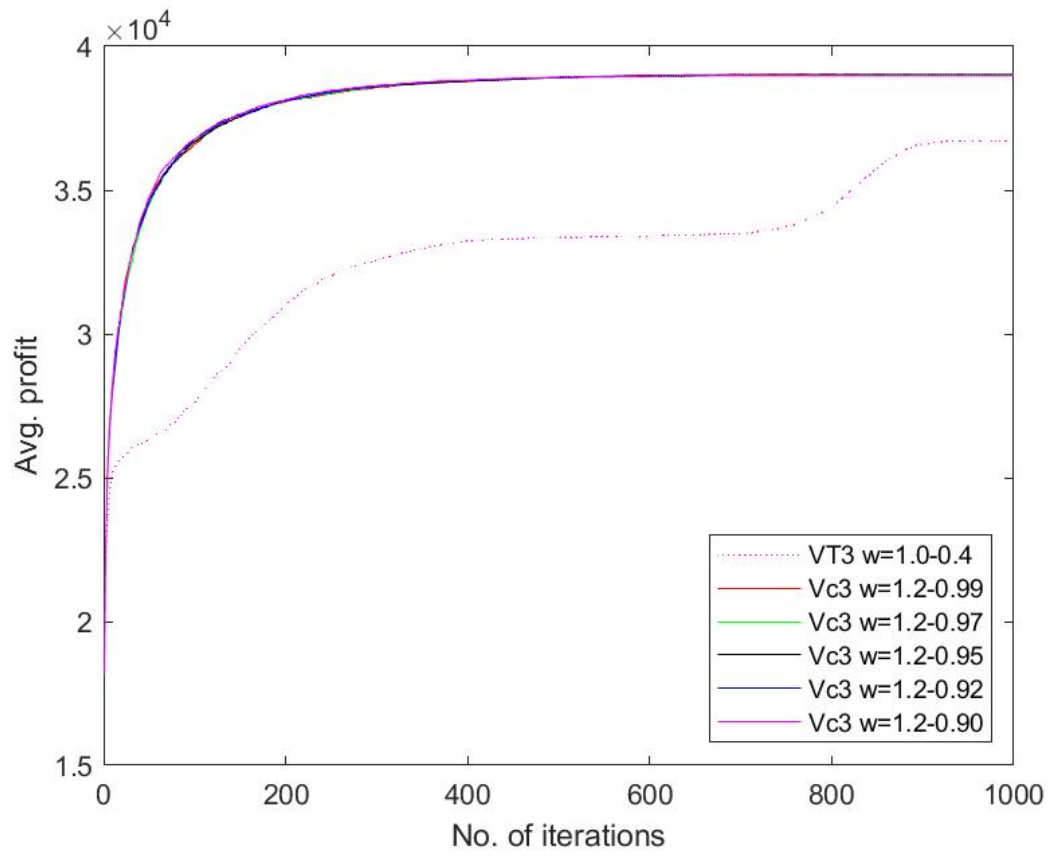

(c) $VC_{V3}$

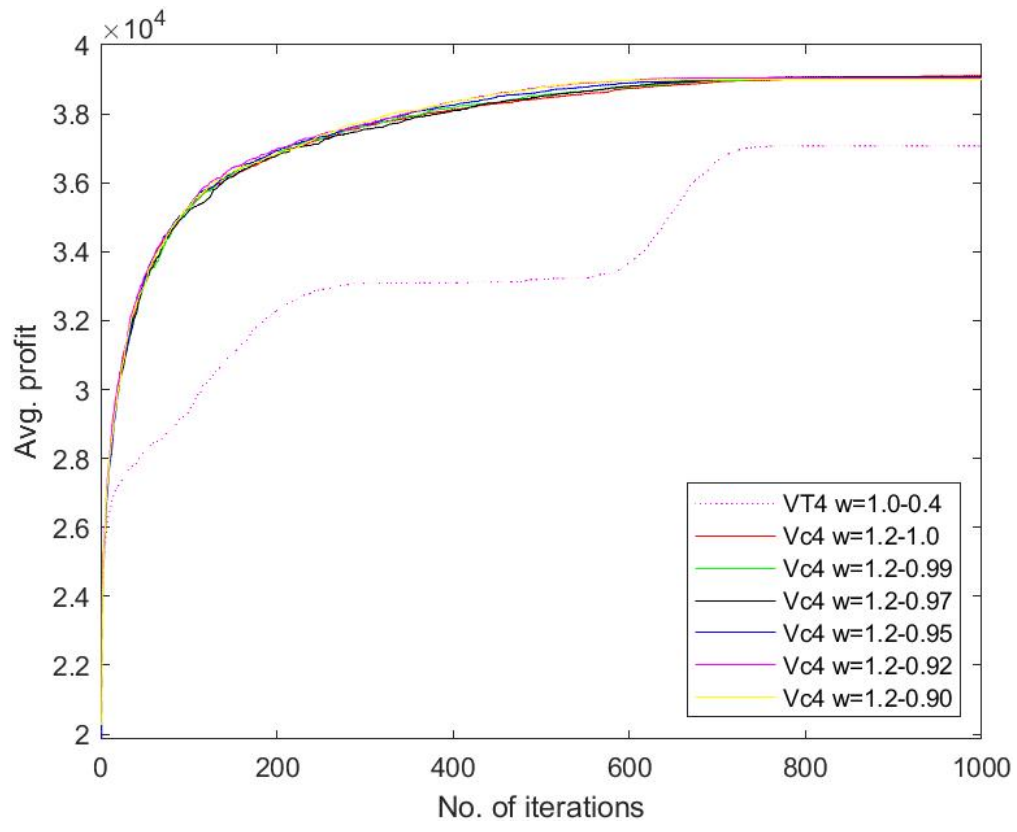

(d) $VC_{V4}$

Fig.6 Comparison of optimal w parameters before and after fine-tuning

Table 5 The effect of fine-tuning the later w value of the improved four algorithms

| | w | Average number of convergence rounds | Averaged optimal profit | Averaged Useless Jump Volume | | w | Average number of convergence rounds | Averaged Optimal profit | Averaged Useless Jump Volume |
|---|---|---|---|---|---|---|---|---|---|
| $VC_{v1}$ | 1.0 | 658.80 | 39238.00 | 82601 | $VC_{v2}$ | 1.0 | 447.10 | 39134.20 | 38006 |
| | 1.0-0.99 | 406.40 | 39215.00 | 53397 | | 1.0-0.99 | 208.60 | 39072.70 | 24778 |
| | 1.0-0.97 | 321.40 | 39156.95 | 40294 | | 1.0-0.97 | 184.25 | 38991.80 | 19514 |
| | 1.0-0.95 | 276.35 | 39136.40 | 36273 | | 1.0-0.95 | 164.25 | 38886.40 | 17467 |
| $V_{T1}$ | 0.6 | 596.95 | 38684.55 | 83367 | $V_{T2}$ | 1.0-0.4 | 693.45 | 37446.85 | 164811 |
| $VC_{v3}$ | 1.2-0.99 | 769.24 | 39048.61 | 42790 | $VC_{v4}$ | 1.2-0.99 | 875.60 | 39062.10 | 58758 |
| | 1.2-0.97 | 693.40 | 39024.64 | 39312 | | 1.2-0.97 | 805.50 | 39076.26 | 54261 |
| | 1.2-0.95 | 645.04 | 39000.10 | 36327 | | 1.2-0.95 | 740.64 | 39032.14 | 49660 |
| | 1.2-0.92 | 575.15 | 38992.93 | 32952 | | 1.2-0.92 | 668.38 | 39046.22 | 44523 |
| | 1.2-0.90 | 535.73 | 38977.41 | 30824 | | 1.2-0.90 | 620.60 | 38996.86 | 42054 |
| $V_{T3}$ | 1.0-0.4 | 905.99 | 36711.45 | 103803 | $V_{T4}$ | 1.0-0.4 | 725.46 | 37065.62 | 112125 |

From Fig. 6, it can be seen that when fine-tuning the optimal w value in the later stage, from 1.0 to 0.99, 0.97, 0.95 (for VCv1-BPSO, VCv2-BPSO) or from 1.2 to 0.99, 0.97, 0.95, 0.92, 0.90 (for VCv3-BPSO, VCv4-BPSO), the change in the optimal profile value that can be searched for is very small and significantly better than the algorithm before improvement.

From the specific values in Table 5, it can be seen that for VCv1 and VCv2, when the value of w is adjusted from 1.0 to 0.95 in the later stage, the number of convergence rounds decreases by 59.1% and 63.3% respectively, but the optimal profit value only decreases by 0.259% and 0.633%, which is still much higher than that before improvement. At the same time, the average Useless Jump Volume is decreased by 56.1% and 54.0%, respectively. Similarly, as shown in Table 7, for VCv3 and VCv4, when the value of w is adjusted from the optimal 0.99 to 0.90 in the later stage, the number of convergence rounds decreases by 30.4% and 29.1%, respectively, but the optimal profit value only decreases by 0.182% and 0.167%. At the same time, the average number of Useless Jump Volume decreased by 28.0% and 29.1%, respectively.

From the reduction of Useless Jump Volume, it can also be seen that by fine-tuning the w value in the later stage, the repetitive space search volume can be significantly reduced, thereby greatly improving the convergence efficiency of the algorithm.

#### 5.2.4 Experiments of the weakly correlated instances (WCI) and the strongly correlated instances (SCI)

For the 100 dimensional WCI (the weakly correlated instances) and SCI (the strongly correlated instances) test cases, experiments were conducted on the 4 algorithms before and after improvement. The w parameter was tested in multiple sets according to the values of the previous experiments, and the optimal w value for each algorithm was selected for comparative experiments. The optimal w values for each algorithm after testing are shown in Table 6.

Table 6 The optimal w obtained from SCI 100 and UCI 100 experiments under four methods

| Instance | $V_T$ | optimal w | VCv | optimal w |
|---|---|---|---|---|
| WCI 100 | $V_{T1}$ | 0.6 | VCv1 | 1.0 |
| | $V_{T2}$ | 1.0-0.4 | VCv2 | 1.1-0.99 |
| | $V_{T3}$ | 0.6 | VCv3 | 1.2-1.0 |
| | $V_{T4}$ | 1.0-0.4 | VCv4 | 1.2-0.99 |
| SCI 100 | $V_{T1}$ | 0.6 | VCv1 | 1.0 |
| | $V_{T2}$ | 1.0-0.4 | VCv2 | 1.1-0.99 |
| | $V_{T3}$ | 0.6 | VCv3 | 1.2-1.0 |
| | $V_{T4}$ | 1.0-0.4 | VCv4 | 1.2-1.0 |

The experimental results are shown in the following Fig.7. It can be seen that, compared with UCI, the experimental results of SCI and WCI are poorer, and the effect of velocity correction is smaller. The strongly correlated knapsack problem has always been a difficult problem to solve in academia, because its optimal and suboptimal solutions are often concentrated in a very small local space, which is likely to be missed during the search process.

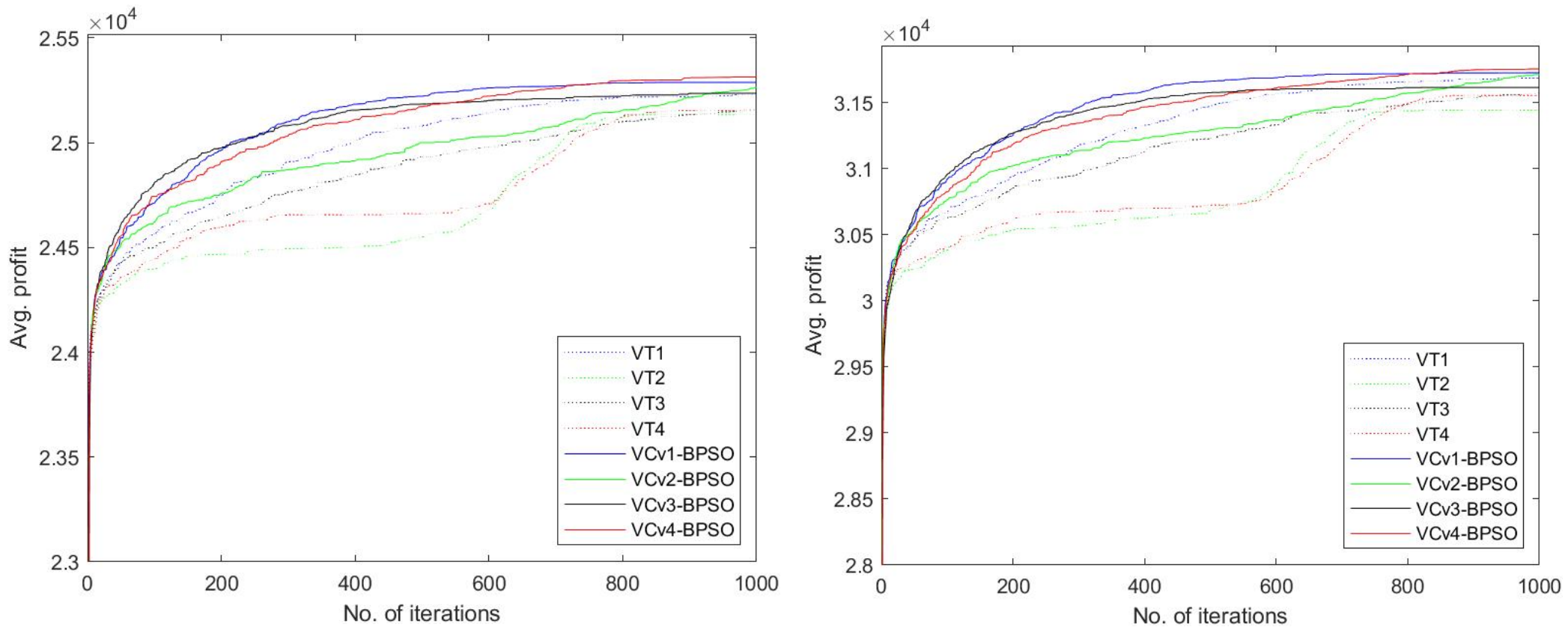

(a) Experiment comparison of WCI 100 (b) Experiment comparison of WCI 100

Fig.7 Experimental results of WCI100 and SCI100 before and after improvement

## 5.3 Comparison of algorithms before and after improvement for medium and high dimensional knapsack

5.3.1 Comparison of maximum profit value found in search

Based on the experience of low dimensional experiments, for test cases with dimensions of 500 and 1000, multiple w parameters were tested for the four V-shaped transformation functions, and the results were compared under the optimal parameters of each algorithm.

Firstly, the most random UCI test cases are selected for experimentation. The experimental parameters before improvement were: w=0.4, w=0.5, w=0.6, w=1.1~0.4, w=1.0~0.4, w=0.9~0.4, w=1.1~0.45, w=1.0~0.45, w=0.9~0.45 (most of which do not converge when w exceeds 0.4 in the later stage); The improved experimental parameters are: w=1.0, w=0.99, w=1.2~0.99, w=1.2~0.97, w=1.2~0.95, w=1.2~0.92, w=1.1~0.99, w=1.1~0.97, w=1.1~0.95, w=1.1~0.92, w=1.0~0.99, w=1.0~0.97, w=1.0~0.92. The comparison of results under the optimal parameters before and after improvement is shown in Table 7.

Table 7 Experiment Comparison for 500 and 1000 dimensional knapsacks

| Test Case | $V_T$ | Optimal w | Ratio of the searched max profit to the true max profit | $VC_v$ | Optimal w | Ratio of the searched max profit to the true max profit |
|---|---|---|---|---|---|---|
| UCI 500 | $V_{T1}$ | 0.6 | 92.7% | $VC_{v1}$ | 1.0-0.99 | **99.4%** |
| | $V_{T2}$ | 0.9-0.4 | 88.8% | $VC_{v2}$ | 1.0-0.99 | **98.8%** |
| | $V_{T3}$ | 0.6 | 91.6% | $VC_{v3}$ | 1.1-0.99 | **99.1%** |
| | $V_{T4}$ | 0.9-0.4 | 90.7% | $VC_{v4}$ | 1.1-0.99 | **99.1%** |
| UCI 1000 | $V_{T1}$ | 0.6 | 88.3% | $VC_{v1}$ | 1.0-0.99 | **98.2%** |
| | $V_{T2}$ | 0.9-0.4 | 83.7% | $VC_{v2}$ | 1.0-0.99 | **97.5%** |
| | $V_{T3}$ | 0.6 | 87.3% | $VC_{v3}$ | 1.1-0.99 | **98.2%** |
| | $V_{T4}$ | 0.9-0.4 | 85.5% | $VC_{v4}$ | 1.0 | **98.7%** |

As shown in Table 7, for both 500 dimensional and 1000 dimensional backpacks, and no matter which conversion function is used, the speed correction method proposed in this article has achieved high optimization results. For the 500 dimensional knapsack, the four BPSO algorithms improved by an average of 8.15% by correcting the velocity legacy term; For 1000 dimensional backpacks, the four BPSO algorithms improved by an average of 11.95%. Moreover, compared with the low dimensional backpack experiment in section 5.2.1, the velocity legacy term correction has a more significant improvement on the mid dimensional and high-dimensional backpack test cases, and the higher the dimension, the more significant the optimization.

Moreover, through the above table, we can also observe that after adjusting for the velocity legacy term, the differences among the four conversion functions have decreased, and the final optimal profit value tends to be consistent. For the 500 dimensional knapsack, the ratio of the optimal solution to the actual maximum profit value is around 99%, while for the 1000 dimensional knapsack, this ratio is around 98%. They once again prove that the effect of velocity legacy term correction is greater than the effect of different transformation function selection.

From Figures 8 and 9, it can be seen that the effect is basically consistent with the low dimensional knapsack experiment. The modified algorithm can quickly search for better solutions in the early stage of iteration and quickly enter a gradually converging state. The convergence speed and the final optimal solution found are significantly better than the previous algorithms. It can be seen that the improvement method of the velocity legacy term has good adaptability to both medium and high-dimensional backpacks.

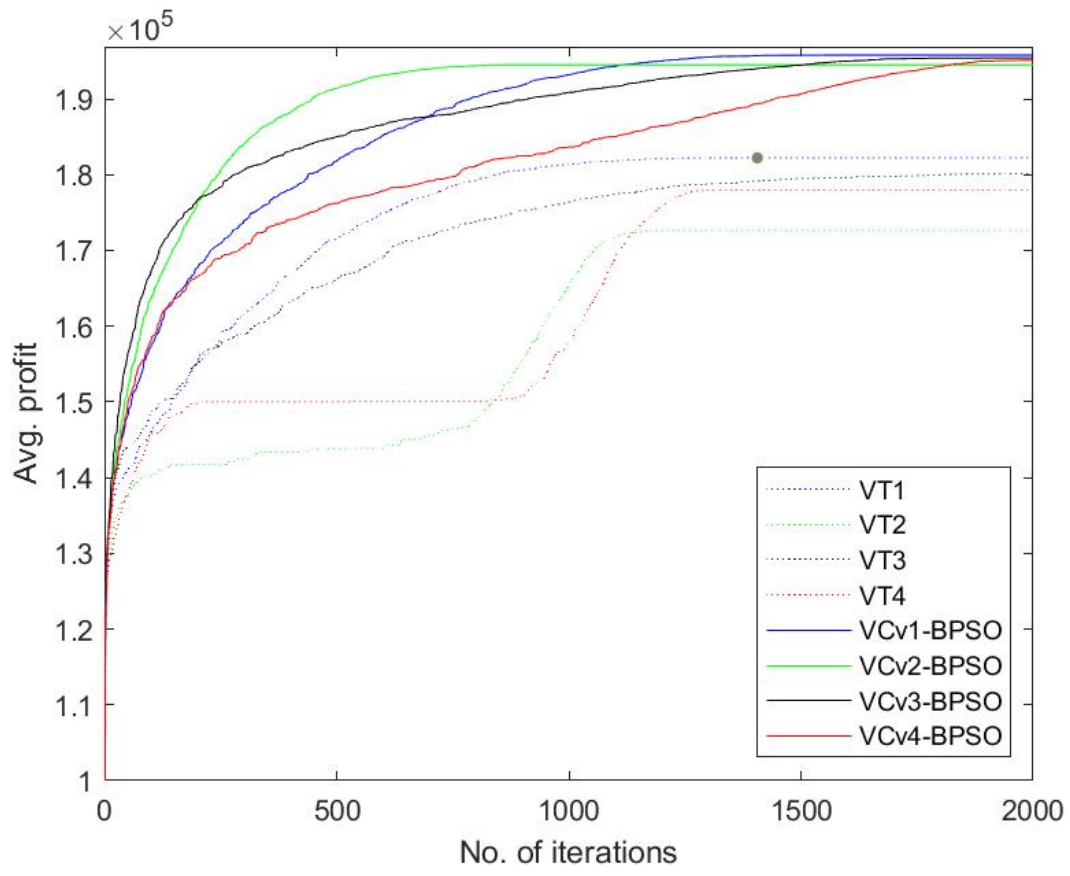

Fig.8 Experimental comparison of algorithms before and after improvement for UCI 500

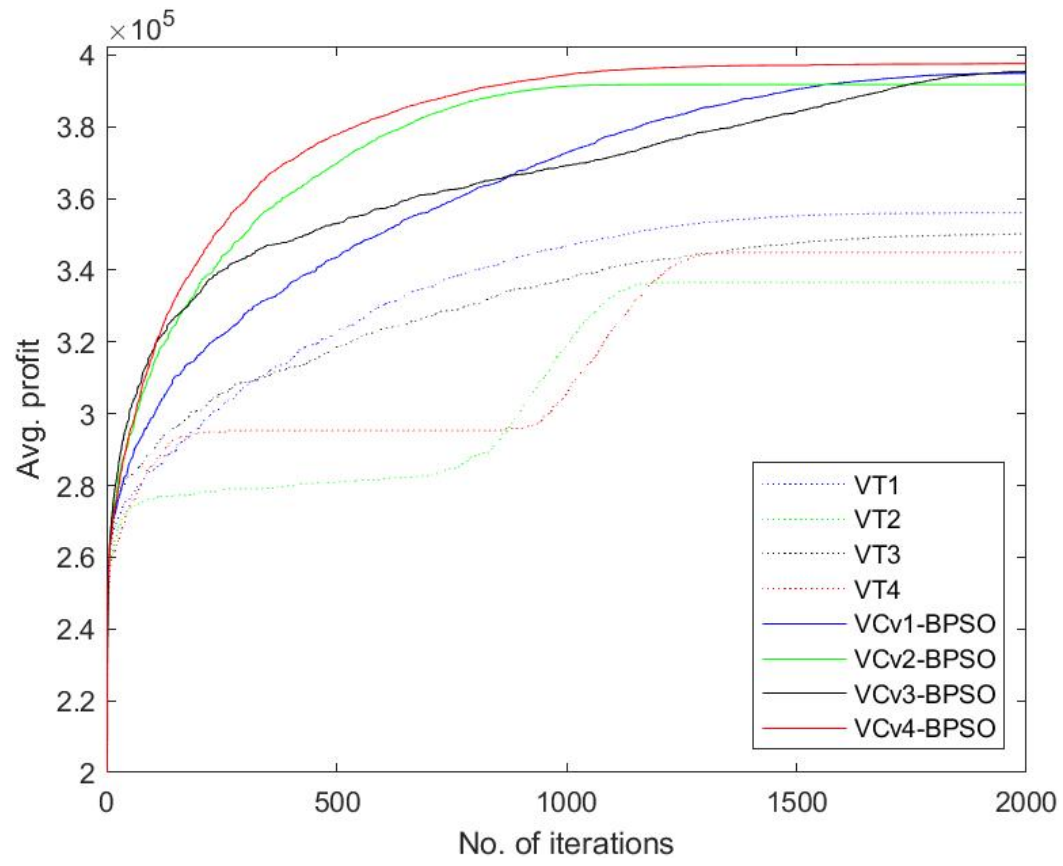

Fig.9 Experimental comparison of algorithms before and after improvement for UCI 1000

## 5.4 Experiments and analysis based on Effective Hamming distance and Useless Jump Volume

Below, four BPSO algorithms will be tested using the two indicators of Effective Hamming distance proposed in this article and Hamming distance in [26] to further analyze the real reasons for the effectiveness of the improved method.

We compare the changes in Hamming distance (Dist) and effective Hamming distance (Dist_eff) of four BPSO methods before and after improvement with increasing iteration times. Taking the 100 dimensional UCI backpacks as the test case, with an initial particle count of 20, the average result after 20 executions is shown in Figure 10. The algorithms before and after improvement have selected the optimal w value.

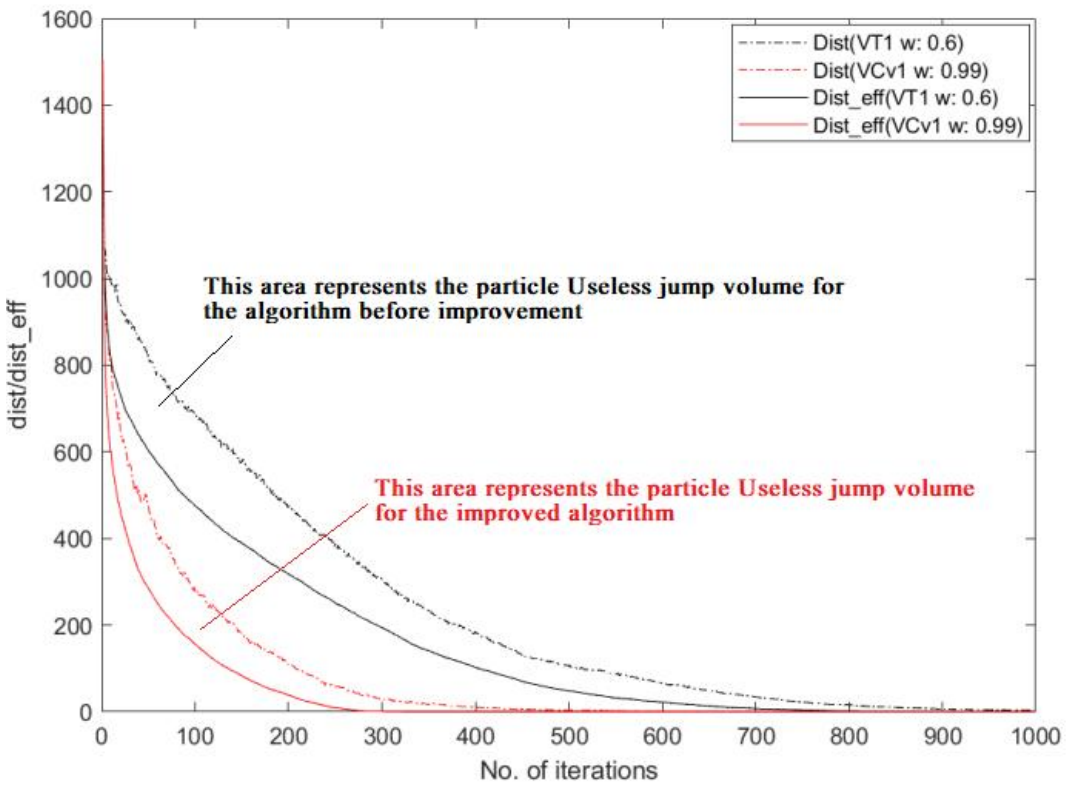

(a) Useless jump volume comparison between VT1 and $VC_{V1}$

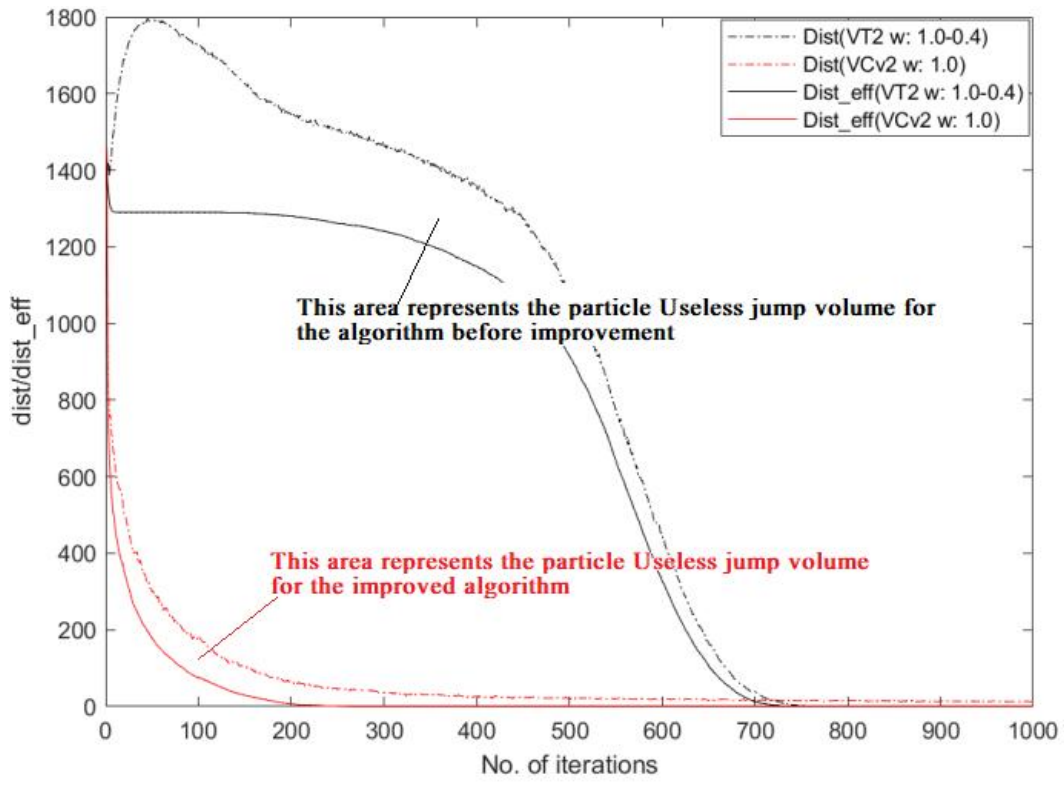

(b) Useless jump volume comparison between VT2 and $VC_{V2}$

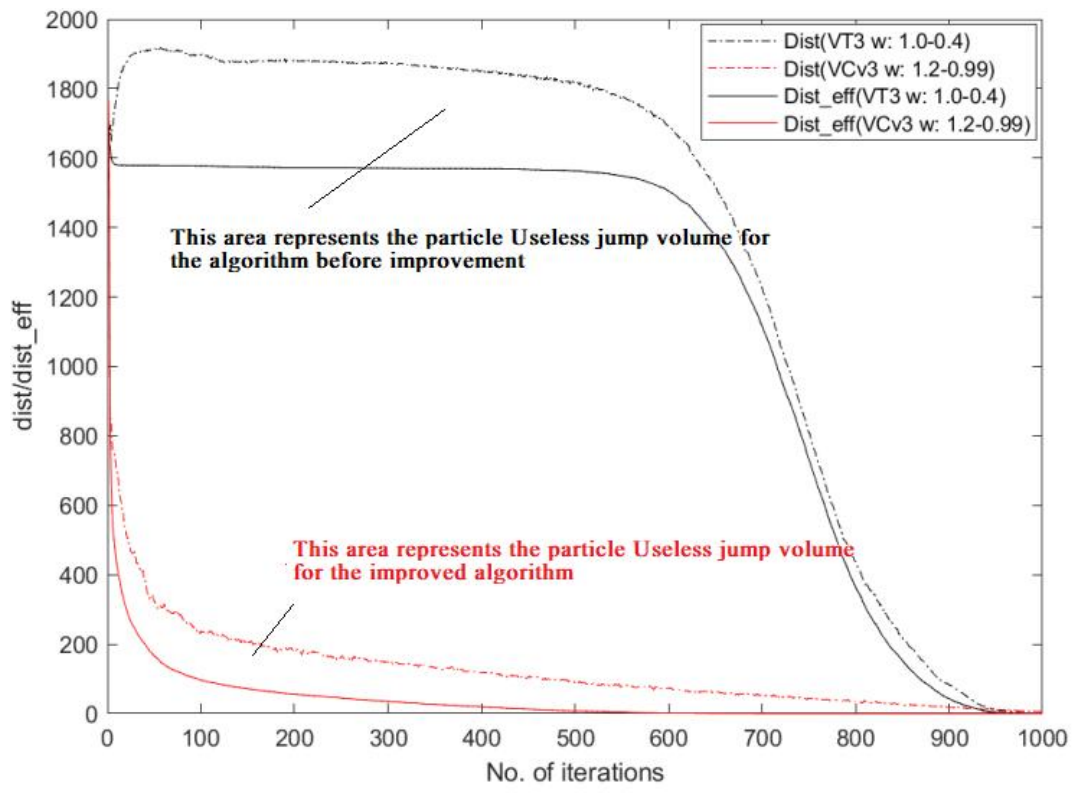

(b) Useless jump volume comparison between VT3 and VCV3

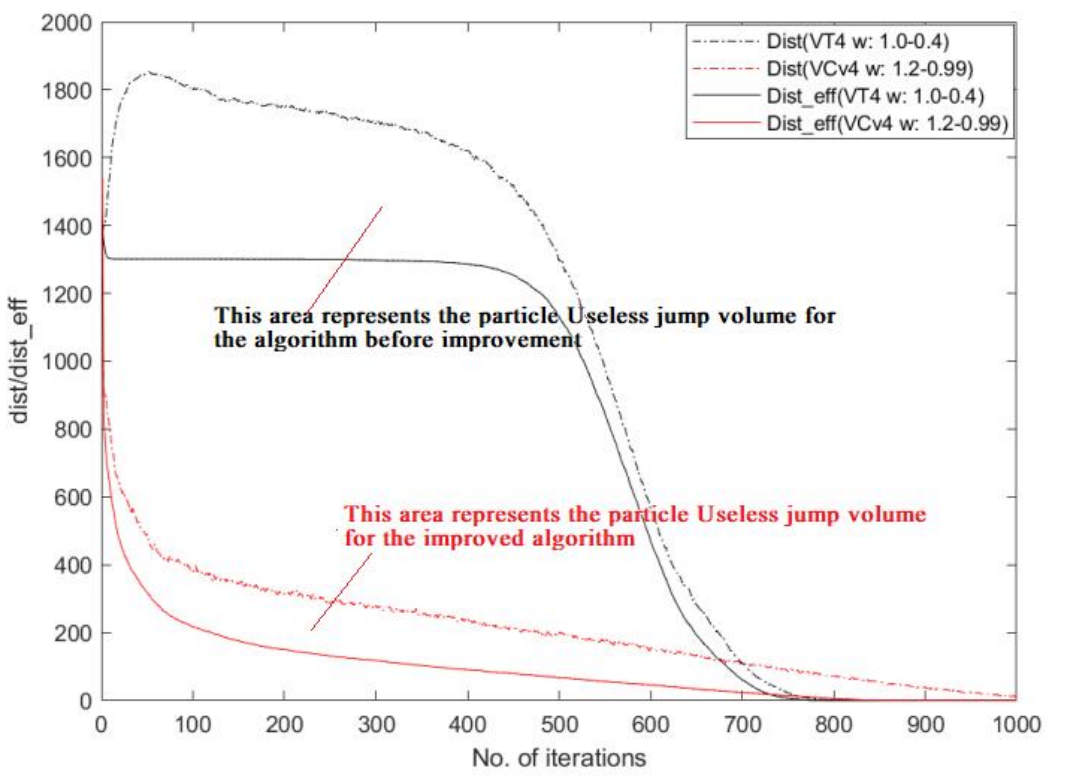

(d) Useless jump volume comparison between VT4 and VCV4

Fig.10 Useless jump volume comparison before and after improvement

According to the explanation in Section 4.2, the effective Hamming distance gain proposed in this article reflects the algorithm's ability to explore new spaces, while the Hamming distance characterizes the activity of particles. But being active does not mean conducting an effective search, and the difference between the two indicators reflects the amount of meaningless jumps of particles in the repeating space. From the figures (b), (c), and (d), it is evident that the effective Hamming distance gain of the algorithm without speed correction, represented by the black solid line, remains high for a long time. This indicates that the algorithm has strong search ability. However, in the later stage of the algorithm, the effective Hamming distance gain suddenly decreases rapidly with a steep slope, which also makes the original algorithm prone to falling into local optima.

The four improved algorithms have a relatively smooth effective Hamming distance gain decrease curve (red solid line) throughout the entire lifecycle. In the early stage of the algorithm, due to the correction of speed, particle behavior is more rational and purposeful, so compared with the previous algorithm, it avoids many blind searches for unnecessary space. In the later stage of execution, the decline curve of exploitation ability is also relatively smooth, which is why the improved algorithm is less likely to fall into local optima.

On the other hand, the area between the two black lines and the two red lines respectively represents the amount of useless jumps in the repeating space of the algorithms before and after improvement. Obviously, the area of the improved algorithm is smaller, which is also the reason why the improved algorithm is more efficient: the correction of the velocity legacy term significantly reduces the amount of meaningless jumps in the repeating space.

Figure 11 illustrates the changes in the values of the Dist_eff and Dist indicators in the early and late stages of the V-BPSO (without velocity correction) and VCv-BPSO (after velocity correction).

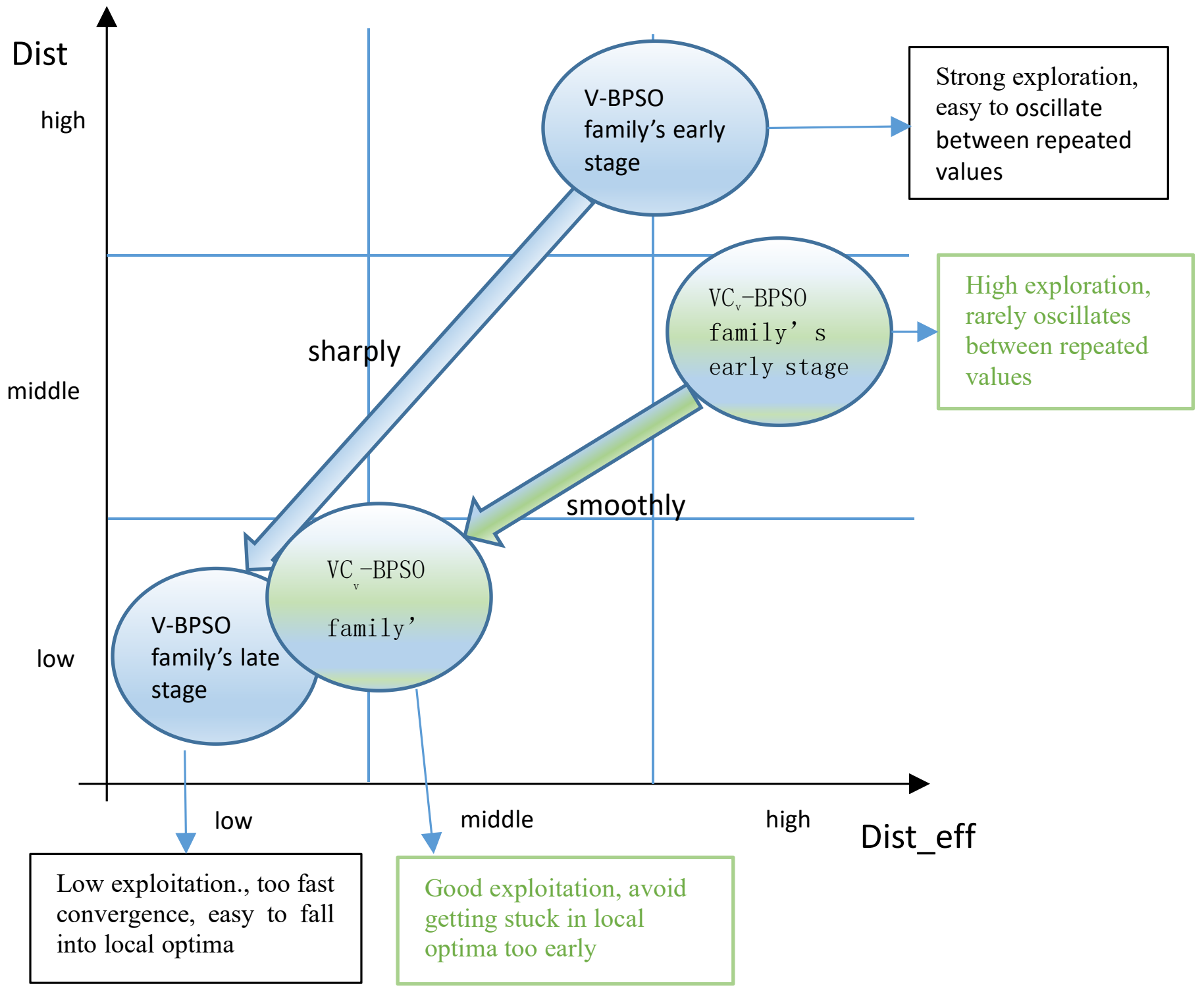

Fig.11 Comparison of Dist_eff and Dist in the whole life cycle of V-BPSOand VCv-BPSO

## 6 Conclusions and prospects

The BPSO algorithm plays an important role in the optimization field of multi-dimensional non convex discrete spaces, and belongs to the field of Swarm Intelligence. And it has important application value in the structural optimization and parameter optimization of deep artificial networks. This article aims to improve the common problems of weak exploitation ability and easy falling into local optima in the later stage of BPSO and its various variants. Taking the V-Shaped BPSO family, which has the best performance in BPSO field, as the research object, it is found that its speed update function generally has obvious shortcomings when inheriting the speed update function of PSO algorithm. The calculation method of the velocity legacy term contained in it is incorrect, because it does not conform to the basic idea of the original PSO algorithm. The existence of this problem directly leads to all V-Shaped BPSO family algorithms being prone to repeated oscillations in the repetition space during the search process. In the later stage, the algorithm has to rely on reducing the w value to force convergence, which further leads to the algorithm losing its exploration ability too quickly and easily falling into the optimal solution. Because various studies have shown that the V-shaped BPSO family is currently the best performing and most promising type of BPSO algorithm, improving the V-Shaped BPSO algorithm is of great significance. An improved method for velocity update formula based on velocity residual term correction is proposed, which makes the improved formula more in line with the basic principles of the original PSO. Taking the the four most commonly used V-shaped transformation functions of BPSO as examples, calculate their respective velocity correction formulas, and test the four improved BPSO algorithms using the 01 knapsack problem in 100, 500, and 1000 dimensions as examples. The experimental results indicate that regardless of which commonly used V-Shaped conversion function is chosen, the effect of velocity correction is significant, and the higher the dimension, the more significant the improvement, with its effect much greater than the selection effect of different conversion functions. At the same time, the search efficiency of the algorithm has also been greatly improved, and the algorithm can naturally converge without relying on the adjustment of parameters such as w, Vmax, Vmin, etc, and does not require difficult leveling between exploration and exploitation. This also breaks the traditional belief that the key to studying BPSO algorithms lies in the selection of transformation functions and parameter tuning. Therefore, this article believes that the proposal of the BPSO algorithm based on velocity residual term correction is an important breakthrough in the field of Swarm Intelligence, which is expected to help various application scenarios of multi-dimensional discrete space optimization problems find better solutions in less time.

However, it should also be noted that further improvement is needed in the subsequent research of this article. The current improved algorithm performs poorly on Strongly correlated instance (SCI). The strongly correlated knapsack problem has always been a difficult problem in the field of optimization[27], as its optimal and suboptimal solutions are concentrated in a small area of multidimensional space, which is easily overlooked during the search process. We will continue to study this part in the future, which can be improved by increasing the initial population size or separating the space for multi head local search in the initial stage.

# Appendix: Derivation of v' for $VC_{v1}$-BPSO, $VC_{v3}$-BPSO and $VC_{v4}$-BPSO

For VCv1-BPSO, VCv3-BPSO and VCv4-BPSO,the velocity correction function can be deduced as following.

(1) Transfer function 1: $sigm(v_{id}) = \left|\frac{2}{\pi}\arctan(\frac{\pi}{2}v_{id})\right|$

According to Equation(3), $sigm(v') = 1 - sigm(v)$

$$sigm(v) = \left|\frac{2}{\pi}\arctan(\frac{\pi}{2}v)\right|$$

$1^\circ$ if $v > 0$,

$$\arctan(\frac{\pi}{2}v) \in \left[0, \frac{\pi}{2}\right), sigm(v) \in [0,1)$$

$$\therefore sigm(v) = \frac{2}{\pi}\arctan(\frac{\pi}{2}v)$$

And by Equation(3), $sigm(v') \in (0,1]$

$$\therefore sigm(v') = \frac{2}{\pi}\arctan\left(\frac{\pi}{2}v'\right)$$

$$\therefore \frac{2}{\pi}\arctan(\frac{\pi}{2}v') = 1 - \frac{2}{\pi}\arctan\left(\frac{\pi}{2}v\right)$$

$$\Rightarrow \frac{2}{\pi}\left[\arctan(\frac{\pi}{2}v') + \arctan(\frac{\pi}{2}v)\right] = 1$$

$$\Rightarrow \arctan(\frac{\frac{\pi}{2}v' + \frac{\pi}{2}v}{1 - \frac{\pi^2}{4}v'v}) = \frac{\pi}{2}$$

$$\therefore \frac{\frac{\pi}{2}v' + \frac{\pi}{2}v}{1 - \frac{\pi^2}{4}v'v} \to +\infty$$

$$\therefore \frac{\pi^2}{4}v'v = 1$$

$$\therefore v' = \frac{4}{\pi^2 v}$$

$2^\circ$ if $v \le 0$,

$\because v' \le 0$

$$\therefore -\frac{2}{\pi}\arctan\frac{\pi}{2}v' = 1+\frac{2}{\pi}\arctan\frac{\pi}{2}v$$

$$\Rightarrow \arctan\frac{\pi}{2}v' + \arctan\frac{\pi}{2}v = -\frac{\pi}{2}$$

$$\Rightarrow \arctan\left(\frac{\frac{\pi}{2}v' + \frac{\pi}{2}v}{1-\frac{\pi^2}{4}v'v}\right) = -\frac{\pi}{2}$$

$$\Rightarrow \frac{\frac{\pi}{2}v' + \frac{\pi}{2}v}{1-\frac{\pi^2}{4}v'v} \to -\infty$$

$$\because \frac{\pi}{2}v' + \frac{\pi}{2}v \le 0$$

$$\therefore 1-\frac{\pi^2}{4}v'v = 0$$

$$\therefore v' = \frac{4}{\pi^2 v}$$

(2) Transfer function 3: $sigm(v_{id}) = |\tanh(v_{id})|$

$$sigm(v_{id}) = |\tanh(\mathrm{v_{id}})| = \left|\frac{e^v - e^{-v}}{e^v + e^{-v}}\right|$$

$1^\circ$ $v \ge 0$,

$\therefore \tanh(\mathrm{v}) \ge 0$

$$\therefore \mathrm{sigm(v)} = \frac{e^v - e^{-v}}{e^v + e^{-v}}$$

*According* Equation(3), $\frac{e^{v'} - e^{-v'}}{e^{v'} + e^{-v'}} = 1 - \frac{e^v - e^{-v}}{e^v + e^{-v}}$

$$\Rightarrow \frac{e^{v'} - e^{-v'}}{e^{v'} + e^{-v'}} = \frac{2}{e^{2v}+1}$$

$$\Rightarrow \frac{e^{2v'}-1}{e^{2v'}+1} = \frac{2}{e^{2v}+1}$$

$$\Rightarrow e^{2v'} = \frac{3+e^{2v}}{e^{2v}-1}$$

$$\Rightarrow e^{2v'} = \frac{3e^{-v}+e^v}{e^v - e^{-v}}$$

$$\therefore \textit{if } \mathrm{v} \ge 0, v' = \frac{1}{2}\ln\left(\frac{3e^{-v}+e^v}{e^v - e^{-v}}\right)$$

$2^\circ$ $v < 0$

$2^\circ\ v < 0,$

$\therefore \tanh(v) < 0$

$\therefore sigm(v) = -\dfrac{e^v - e^{-v}}{e^v + e^{-v}} = \dfrac{e^{-v} - e^v}{e^v + e^{-v}}$

$According$ Equation(3), $\dfrac{e^{-v'} - e^{v'}}{e^{v'} + e^{-v'}} = 1 - \dfrac{e^{-v} - e^v}{e^v + e^{-v}}$

$\Rightarrow \dfrac{e^{-v'} - e^{v'}}{e^{v'} + e^{-v'}} = \dfrac{e^v + e^{-v} - e^{-v} + e^v}{e^v + e^{-v}} = \dfrac{2e^v}{e^v + e^{-v}}$

$\Rightarrow \dfrac{1 - e^{2v'}}{e^{2v'} + 1} = \dfrac{2e^{2v}}{e^{2v} + 1}$

$\Rightarrow (1 - e^{2v'})(e^{2v} + 1) = 2e^{2v}(e^{2v'} + 1)$

$\Rightarrow 1 - e^{2v} = 3e^{2v}e^{2v'} + e^{2v'}$

$\therefore e^{2v'} = \dfrac{1 - e^{2v}}{3e^{2v} + 1} = \dfrac{e^{-v} - e^v}{3e^v + e^{-v}}$

$\therefore \text{if v} < 0, v' = \dfrac{1}{2}\ln\left(\dfrac{e^{-v} - e^v}{3e^v + e^{-v}}\right)$

## (3) Transfer function 4: $sigm(v_{id}) = \dfrac{2}{1 + e^{-|v_{id}|}} - 1$

Let $x = sigm(v)$,

According to Equation(3),

$1 - x = \dfrac{2}{1 + e^{-|v'|}} - 1$

$\Rightarrow 2 - x = \dfrac{2}{1 + e^{-|v'|}}$

$1 + e^{-|v'|} = \dfrac{x}{2 - x}$

$e^{-|v'|} = \dfrac{x}{2 - x}$

$-|v'| = \ln\dfrac{x}{2 - x}$

$|v'| = \ln\dfrac{x}{2 - x}$

$\therefore$ if $v \geq 0, |v'| = v'$, and if $v < 0, |v'| = -v'$.

$1^\circ\ v \geq 0,$

$x = sigm(v) = \dfrac{2}{1 + e^{-v}} - 1 = \dfrac{1 - e^{-v}}{1 + e^{-v}}$

subsititude in x :

$v' = \ln\left(\dfrac{2 - x}{x}\right) = \ln(\dfrac{2}{x} - 1)$

$= \ln\left(\dfrac{2 + 2e^{-v}}{1 - e^{-v}} - 1\right) = \ln\left(\dfrac{1 + 3e^{-v}}{1 - e^{-v}}\right)$

$2^\circ$ if $v < 0,$

$x = sigm(v) = 1 - \dfrac{2}{1 + e^{-v}} = \dfrac{e^{-v} - 1}{1 + e^{-v}}$

Substidue in $x$ :

$v' = -\ln\left(\dfrac{2 - x}{x}\right) = -\ln(\dfrac{2}{x} - 1)$

$= -\ln\left(\dfrac{2 + 2e^{-v}}{e^{-v} - 1} - 1\right) = -\ln\left(\dfrac{e^{-v} + 3}{e^{-v} - 1}\right)$

$= \ln\left(\dfrac{e^{-v} - 1}{e^{-v} + 3}\right)$